\documentclass[accepted, mathfont=newtx]{uai2025} 
                        
\usepackage[american]{babel}

\usepackage{natbib} 
\usepackage{mathtools} 
\usepackage{tikz} 
\usepackage{mathrsfs}           
\mathtoolsset{showonlyrefs}     
\usepackage{graphicx}           
\usepackage{subcaption}         
\usepackage[utf8]{inputenc} 
\usepackage{booktabs}       
\usepackage{nicefrac}       
\usepackage{microtype}      
\usepackage{xcolor}         
\usepackage[inline]{enumitem}
\usepackage{algorithm}
\usepackage{algpseudocode}
\usepackage{svg}
\usepackage{dsfont}

\usepackage{amsthm}
\usepackage{thmtools, thm-restate}

\usepackage[skip=2pt]{caption}

\usepackage{hyperref}
\hypersetup{
  colorlinks   = true,    
  urlcolor     = blue,    
  linkcolor    = red,    
  citecolor    = blue      
}

\title{Distributional Reinforcement Learning with \\Dual Expectile-Quantile Regression}

\author[* 1]{\href{mailto:<s.jullien@uva.nl>?Subject=Your UAI 2025 paper}{\textcolor{black}{Sami Jullien}}}
\author[* 1,2]{\href{mailto:<romain.deffayet@naverlabs.com>?Subject=Your UAI 2025 paper}{\textcolor{black}{Romain Deffayet}}}
\author[1]{\href{mailto:<p.t.groth@uva.nl>?Subject=Your UAI 2025 paper}{\textcolor{black}{Paul Groth}}}
\author[2]{\href{mailto:<jean-michel.renders@naverlabs.com>?Subject=Your UAI 2025 paper}{\textcolor{black}{Jean-Michel Renders}}}
\author[1]{\href{mailto:<m.derijke@uva.nl>?Subject=Your UAI 2025 paper}{\textcolor{black}{Maarten de Rijke}}}
\affil[1]{%
    University of Amsterdam \\
    Amsterdam, The Netherlands
}
\affil[2]{%
    Naver Labs Europe\\
    Meylan, France
  }
  
\begin{document}

\maketitle
\def\thefootnote{*}\footnotetext{Both authors contributed equally to the paper.}

\begin{abstract}
    Distributional reinforcement learning (RL) has proven useful in multiple benchmarks as it enables approximating the full distribution of returns and extracts rich feedback from environment samples. The commonly used quantile regression approach to distributional RL -- based on asymmetric $L_1$ losses -- provides a flexible and effective way of learning arbitrary return distributions. In practice, it is often improved by using a more efficient, asymmetric hybrid $L_1$-$L_2$ Huber loss for quantile regression.
    However, by doing so, distributional estimation guarantees vanish, and we empirically observe that the estimated distribution rapidly collapses to its mean. Indeed, asymmetric $L_2$ losses, corresponding to expectile regression, cannot be readily used for distributional temporal difference learning. 
    
    Motivated by the efficiency of $L_2$-based learning, we propose to jointly learn expectiles and quantiles of the return distribution in a way that allows efficient learning while keeping an estimate of the full distribution of returns. We prove that our proposed operator converges to the distributional Bellman operator in the limit of infinite estimated quantile and expectile fractions, and we benchmark a practical implementation on a toy example and at scale. On the Atari benchmark, our approach matches the performance of the Huber-based IQN-1 baseline after $200$M training frames but avoids distributional collapse and keeps estimates of the full distribution of returns. Code: \url{https://github.com/samijullien/ieqn}
\end{abstract}

\section{Introduction}
\label{sec:intro}
Distributional reinforcement learning (RL)~\citep{distributional-book} aims to maintain an estimate of the full distribution of expected returns rather than only the mean. Compared to a mean-based approach, it can be used to better capture the uncertainty in the transition matrix of the environment~\citep{c51}, as well as the stochasticity of the policy being evaluated, which may enable faster and more stable training by making better use of the data samples~\citep{mavrin2019distributional}.

Non-parametric approximations of the return distribution learned by quantile regression have proven to be very effective in several domains~\citep{iqn, qr-dqn, fqf}, when combined with deep RL agents such as deep Q-networks (DQN)~\citep{dqn} or soft actor-critic (SAC)~\citep{sac}. They come with the major advantage of providing guarantees for the convergence of distributional policy estimation~\citep{qr-dqn}, and in certain cases, of convergence to the optimal policy~\citep{rowland2023analysis}, all while requiring few assumptions on the shape of the return distribution and demonstrating strong empirical performance~\citep{iqn, fqf}. However, the best-performing quantile-based agents are often obtained by replacing the original quantile regression loss function, i.e., an asymmetric $L_1$ loss, by an asymmetric Huber loss, i.e., a hybrid $L_1$-$L_2$ loss. By doing so, distributional guarantees vanish, as the proofs proposed in previous work relied on the $L_1$-based quantile regression~\citep{distributional-book, qr-dqn}. Critically, we show in Section~\ref{sec:atari} that the estimated distributions collapse to their mean in practice. In this paper, \textit{we propose a different approach, based on both quantile and expectile regression, that matches the performance of Huber-based agents while preserving distributional estimation guarantees and avoiding distributional collapse in practice.}

We are not the first to note that asymmetric $L_2$ losses, i.e., that regress \textit{expectiles} of the target distribution, tend to yield degenerate estimated distributions when training agents with temporal difference learning. \citet{er-dqn} note that expectiles of a distribution cannot be interpreted as samples from this distribution, and therefore expectiles other than the mean cannot be directly used to compute the target values in distributional temporal difference learning. Instead, they propose to generate samples from expectiles of the distribution by adding an imputation step, that requires solving a costly root-finding problem. While theoretically justified, we found this approach to be extremely slow in practice, preventing widespread use at scale. In contrast, our dual approach tackles this problem through learning, and only requires an additional two-layer neural network with the computation of a quantile loss function on top of the expectile loss function. This approach therefore adds close to no computational overheads when training Atari agents on modern GPUs.

Our contributions can be summarized as follows:
\begin{itemize}[nosep]
    \item We propose a novel dual expectile-quantile approach to distributional dynamic programming that provably converges to the true value distribution in the limit  of infinite estimated quantile and expectile fractions. 
    \item We release implicit expectile-quantile networks (IEQN),\footnote{Available at \url{https://github.com/samijullien/ieqn}.} a practical implementation of our dual approach based on implicit quantile networks~\citep{iqn}.
    \item We show both on a toy example and at scale on the Atari-5 benchmark that IEQN \begin{enumerate*}[label=(\roman*)]
        \item avoids distributional collapse, and
        \item matches the performance of the Huber-based IQN-1 approach.
    \end{enumerate*}
\end{itemize}
\section{Background}
\label{sec:background}

\subsection{Distributional reinforcement learning}
\label{sec:drl}

We consider an environment modeled by a Markov decision process (MDP) $(\mathcal{S}, \mathcal{A}, R, T, \gamma)$, where $\mathcal{S}$ and $\mathcal{A}$ are a state and action space, respectively, $R(s,a)$ denotes the stochastic reward obtained by taking action $a$ in state $s$, $T(\cdot \mid s,a)$ is the probability distribution over possible next states after taking $a$ in $s$, and $\gamma$ is a discount factor. Furthermore, we write $\pi(\cdot \mid s)$ for a (potentially stochastic) policy selecting the action depending on the current state. 

We consider the problem of finding a policy maximizing the average discounted return, i.e., 
\begin{equation}
    \pi^* = \arg\max_\pi \mathbb{E} \left[\sum_{t=0}^\infty \gamma^t R(s_t, a_t) \right],
\end{equation}
where $a_t \sim \pi(\cdot \mid s_t)$ and $s_{t+1} \sim T(\cdot \mid s_t, a_t)$. We can define the action-value random variable for policy $\pi$ as $Z^\pi : (s,a) \mapsto \sum_{t=0}^\infty \gamma^t R(s_t, a_t)$, with $s_0=s, a_0=a$. We will refer to action-value variables and their estimators as $Z$-functions in the remainder. Note that the $Q$-function, as usually defined in RL~\citep{sutton2018reinforcement}, is given by $Q^\pi(s,a) = \mathbb{E}\left[ Z^\pi(s,a) \right]$. In this work, we consider approaches that evaluate policies through distributional dynamic programming, i.e., by repeatedly applying the distributional Bellman operator $\mathcal{T}^\pi$ to a candidate $Z$-function:
\begin{equation}
    \mathcal{T}^\pi Z(s_t,a_t) = R(s_t,a_t) + \gamma \mathbb{E}_\pi\left[Z(s_{t+1}^\pi,a_{t+1}^\pi)\right].
\label{eq:ddp}
\end{equation}
This operator has been shown to be a contraction in the $p$-Wasserstein distance and therefore admits a unique fixed point $Z^\pi$~\citep{c51}. A major challenge of distributional RL resides in the choice of representation for the action-value distribution, as well as the empirical implementation of the distributional Bellman operator. For simplicity, in the remainder and in line with previous work, we only consider empirical distributions~\citep[Definition~5.5]{distributional-book} (i.e., whose representation can fit in finite memory), and refer to the empirical representation distributional Bellman operator~\citep[Algorithm~5.1]{distributional-book} as $\mathcal{T}^\pi$.

\subsection{Quantile and expectile regression}
\label{section:regression}

Let $Z$ be a real-valued probability distribution. The $\alpha$-\emph{quantile} $q_\alpha$ of $Z$ is defined as a value splitting the probability mass of $Z$ in two parts of weights $\alpha$ and $1 - \alpha$, respectively:
\begin{equation}
    P(z \leq q_\alpha) = \alpha.
\label{quantile-def}
\end{equation}
Therefore, the \textit{quantile function} $Q_Z : \alpha \mapsto q_\alpha $ is the inverse cumulative distribution function: $Q_Z = F_Z^{-1}$. Alternatively, quantiles are given by the minimizer of an asymmetric $L_1$ loss:
\begin{equation}
\mbox{}\hspace*{-1mm}
    q_\alpha = \arg\min_{q} \mathbb{E}_{z \sim Z} \left[\left(\alpha \mathds{1}_{z > q} + (1 - \alpha) \mathds{1}_{z \leq q} \right)\left| z - q \right| \right].
\hspace*{-1mm}\mbox{}    
\label{quantile-reg}
\end{equation}
Expectiles and the \textit{expectile function} $E_Z : \tau \mapsto e_\tau$ are defined analogously, as the $\tau$-expectile $e_\tau$ minimizes the asymmetric $L_2$ loss:
\begin{equation}
\label{expectile-reg}
\mbox{}\hspace*{-1mm}
    e_\tau \!=\! \arg\min_{e} \mathbb{E}_{z \sim Z} \!\left[\left(\tau \mathds{1}_{z > e} + (1 - \tau) \mathds{1}_{z \leq e} \right) \left( z - e \right)^2 \right]\!.
\hspace*{-1mm}\mbox{}    
\end{equation}

\subsection{Quantiles and expectiles in distributional RL}

Quantile regression has been used for distributional RL in many previous studies~\citep[see, e.g.,][]{iqn, qr-dqn, fqf} where a parameterized quantile function $Q_Z^\theta(s,a,\alpha)$ is trained using a quantile temporal difference loss function derived from Eq.~\eqref{quantile-reg}, i.e., for $N$ estimated quantiles:
\begin{equation}
\label{qtd-loss}
\mathcal{L}_Q\left(Q_Z^\theta(s,a, \cdot), \mathbf{z} \right) \!=\! \sum_{i=1}^N \sum_{j=1}^N l_Q(q_i, z_j), 
\end{equation}
with $l_Q(q_i, z_j) \!=\! (\alpha_i \mathds{1}_{z_j > q_i} + (1 - \alpha_i) \mathds{1}_{z_j \leq q_i} )| z_j - q_i |$,
where the trainable quantile values $q_i = Q_Z^\theta(s,a, \alpha_i)$ are obtained by querying the quantile function at various quantile fractions $\alpha_i$, which can be either fixed by the designer~\citep{qr-dqn}, sampled from a distribution~\citep{iqn}, or learned during training~\citep{fqf}. In quantile-based temporal difference (QTD) learning, the target samples $z_j$ can be obtained by querying the estimated quantile function at the next state-action pair: $z_j = r + \gamma Q_Z^\theta(s',a', \alpha_j)$.\footnote{We can have $a' \sim \pi(\cdot \mid s')$, as in actor-critic algorithms, or $a' = 
\arg\max_a \mathcal{Q}_Z^\theta(s',a, \alpha_j)$ as in Q-learning. This section is agnostic to that choice but we refer to~\citep{distributional-book} for convergence analysis in the latter case.} Indeed, because the true quantile function is the inverse CDF of the action-value distribution, \citet{qr-dqn} and \citet{distributional-book} showed that, among $N$-atoms representations, quantiles at equidistant fractions minimize the $1$-Wasserstein distance with the action-value distribution and that the resulting projected Bellman operator is a contraction mapping in such a distance. \citet{rowland2023analysis} extended these results to prove the convergence of QTD learning under mild assumptions. We refer to these studies for a more detailed convergence analysis.

In contrast, expectile-based temporal difference (ETD) learning does not allow the same training loss as the one given by Eq.~\eqref{qtd-loss}. We first write the generic ETD loss derived from Eq.~\eqref{expectile-reg}:
\begin{equation}
\label{etd-loss}
\mathcal{L}_E\left(E_Z^\theta(s,a, \cdot), \mathbf{z} \right) \!=\! \sum_{i=1}^N \sum_{j=1}^N l_E(e_i, z_j), 
\end{equation}
with 
$l_E(e_i, z_j) \!=\! (\tau_i \mathds{1}_{z_j > e_i} + (1 - \tau_i) \mathds{1}_{z_j \leq e_i} )( z_j - e_i )^2$,
and $e_i = E_Z^\theta(s,a, \tau_i)$. Here, choosing $z_j = r + \gamma E_Z^\theta(s',a', \tau_j)$, analogously to QTD learning and non-distributional TD learning, would cause the update to approximate a different distribution because the expectile function is in general not the inverse CDF of the return distribution, meaning that expectiles cannot be considered as samples from the distribution. \citet{er-dqn} formalized this idea using the concept of \textit{Bellman-closedness}, i.e., that the projected Bellman operator yields the same statistics whether it is applied to the target distribution or to the implicit distribution given by statistics of the target distribution (i.e., in our case a uniform mixture of diracs with locations given by quantiles or expectiles). 

\section{Related Work}
\label{sec:related-work}

\subsection{Distributional reinforcement learning}
\label{sec:drl-related}

Distributional reinforcement learning has been shown to result in several benefits over a mean-based approach -- by ascribing randomness to the value of a state-action pair, an algorithm can learn more efficiently for close states and actions~\citep{mavrin2019distributional}, as well as capture possible stochasticity in the environment~\citep{sddrl}. Some works also use distributional RL for risk-sensitive control~\citep{pmlr-v139-fei21a,NEURIPS2022_c88a2bd0, NEURIPS2022_d2511dfb}. Multiple families of approaches have emerged.

Estimating a parameterized distribution is a straightforward approach, and has been explored from both Bayesian~\citep{strens2000bayesian, vlassis2012bayesian} and frequentist~\citep{gtdqn} perspectives. However, this usually requires an expensive likelihood computation, as well as making a restrictive assumption on the shape of the return distribution $Z$. For instance, assuming a normal distribution when the actual distribution is heavy-tailed can yield disappointing results.

Thus, approaches based on non-parametric estimation are also used to approximate the distribution. C51~\citep{c51} quantizes the domain where $Z$ has non-zero density (usually in 51 atoms, hence the name), and performs weighted classification on the atoms, by computing the cross-entropy between $Z$ and $\mathcal{T}^\pi Z$. While C51 increases performance over non-distributional RL, it requires the user to manually set the return bounds and is not guaranteed to minimize any $p$-Wasserstein metric with the target return distribution.

Another important non-parametric approach to the estimation of a distribution is quantile regression. Quantile regression relies on the minimization of an asymmetric $L1$ loss. Estimating quantiles allows one to approximate the action-value distribution without relying on a shape assumption. QR-DQN~\citep{qr-dqn} introduced quantile regression as a way to minimize the $1$-Wasserstein metric between $Z$ and $\mathcal{T}^\pi Z$. ER-DQN~\citep{er-dqn} traded the estimation of quantiles for expectiles, at the cost of a potential distribution collapse, which they prevent via a root-finding procedure.
Further, implicit quantile networks (IQN)~\citep{iqn} sample and embed quantile fractions, instead of keeping them fixed, thereby improving performance. Fully parameterized quantile functions (FQF)~\citep{fqf} add another network generating quantiles fractions to be estimated. 
We build on IQN and its expectile counterpart to propose a well-performing, non-collapsing agent.

\subsection{Expectile regression}
\label{sec:expectile-regression}
Expectiles were originally introduced as a family of estimators of \textit{location parameters} for a given distribution, to palliate possible heteroskedasticity of the error terms in regression~\citep{expectilesoriginal, expectile-blue}.

Expectiles can be seen as mean estimators under missing data~\citep{expbible}. Unlike quantiles, they span the entire convex hull of the distribution's support, and on this ensemble, the expectile function is strictly increasing: an expectile fraction is always associated to a unique value. Expectiles have been used in reinforcement learning successfully before~\citep{er-dqn}, but in a way that requires a slow optimization step to achieve satisfactory performance. Moreover, expectile regression is subject to the same crossing issue as quantiles, albeit empirically less so~\citep{exp-quant-david-goliath}.
Expectiles have also been used in offline reinforcement learning to compute a soft maximum over potential outcomes seen in the offline data~\citep{kostrikov2022offline}.

Importantly for our work, it has been shown that under mild assumptions expectile regression is the best linear unbiased estimator of any location parameter within the range of the distribution, which includes any quantile of the distribution~\citep{expectile-blue}. In particular, expectile regression has lower variance than quantile regression for estimating quantiles of the distribution. This theoretical property has been confirmed empirically by~\citet{exp-quant-david-goliath}. These observations encourage us to use expectile regression as a way to estimate quantiles of the value distribution, which we describe in the next section. In contrast to prior works that proposed numerical solutions to the problem of mapping an estimated expectile to its corresponding quantile~\citep{er-dqn, exp-quant-david-goliath}, we propose a learning-based approach to this problem.

\section{Method}
\label{sec:method}

\subsection{Dual training of quantiles and expectiles}
\label{sec:dual-training}

Expectiles have been suggested to be more efficient than quantiles for function approximation~\citep{expectilesoriginal, exp-quant-david-goliath}, but unlike quantiles, they cannot be directly used to generate proper samples of the estimated return distribution ($z_j$ in Eq.~\eqref{etd-loss}), which are required in distributional dynamic programming. \citet{er-dqn} propose an \textit{imputation strategy}, i.e., a way to generate samples of a distribution that matches the current set of estimated expectiles, by solving a convex optimisation problem. In our experiments, we found that applying this imputation strategy tends to drastically increase the runtime (around 25 times slower in our setup), making experimentation with such methods close to impossible for researchers with modest computing resources. In this paper, we propose to learn a functional mapping between expectiles and quantiles and use the predicted quantiles to generate samples. 

We learn a single $Z$-function using expectile regression. Therefore, we have $\forall (s,a) \in \mathcal{S} \times \mathcal{A}, \tau \in [0,1], \; Z_\theta(s,a,\tau) \mathrel{\hat=} E_{Z(s,a)}(\tau)$, where $Z$ is the true $Z$-function we wish to estimate. Then, we note that for non-deterministic $Z(s,a)$, the expectile function at a given state-action pair $E_{Z(s,a)} \in \mathbb{R}^{[0,1]}$ is a strictly increasing and continuous function that spans the entire convex hull of the distribution's support~\citep{german-paper}. Meanwhile, the quantile function $Q_{Z(s,a)} \in \mathbb{R}^{[0,1]}$ spans the distribution's support. As a consequence, every quantile is a single expectile, i.e., there exists a functional mapping from quantile fractions to expectile fractions. In this work, we propose to learn such a mapper $m_\phi(s,a,\tau) \mathrel{\hat=} E^{-1}_{Z(s,a)} \circ F^{-1}_{Z(s,a)} (\tau) $
using the quantile regression loss function from Eq.~\eqref{qtd-loss}. We then have $\forall (s,a) \in \mathcal{S} \times \mathcal{A}, \tau \in [0,1], \; Z_\theta(s,a,m_\phi(s, a, \tau)) \mathrel{\hat=} Q_{Z(s,a)}(\tau)$. We can then simply query our estimator of quantiles at the next state-action pair to yield a sound imputation step, while the parameters of the $Z$-function are learned through expectile regression. 

For any tuple $(s, a, s', a')$, our proposed update step can be described as follows:
\begin{enumerate}[nosep]
    \item Sample fractions $\hat{\tau} \sim~\mathcal{U}(0,1)$.
    \item Generate approximate samples of the target distribution using the quantile representation: 
    $$\hat{z} = R(s,a) + \gamma Z_\theta(s',a',m_\phi(s', a', \hat{\tau})).$$
    \item Use expectile regression to learn the $Z$-function: 
    $$ Z_\theta(s,a,\hat{\tau}) \leftarrow \min_\theta \mathcal{L}_E\left(Z_\theta(s,a,\hat{\tau}), \hat{z} \right).$$
    \item Use quantile regression to learn the mapper:
    $$m_\phi(s,a,\hat{\tau}) \leftarrow \min_\phi \mathcal{L}_Q\left(Z_\theta(s,a,m_\phi(s, a, \hat{\tau})), \hat{z} \right).$$
\end{enumerate}

\begin{algorithm*}[!ht]
    \caption{Implicit expectile-quantile networks (IEQN) update}
    \label{alg-ieqn}
    \begin{algorithmic}
        \Require $Z$-function $Z_\theta$, mapper $m_\phi$, fractions $(\tau_i)_{i = 1, \dots, N} \sim \mathcal{U}([0,1])$, learning rate $\lambda$.
        \State Collect experience $(s,a,r,s')$
        \For{$i = 1, \dots, N$}
            \State Compute expectile values $e_i \leftarrow Z_\theta(s,a, \tau_i)$ and quantile values $q_i \leftarrow Z_\theta(s,a, m_\phi(\tau_i))$
            \State Compute the greedy next-action: $$a' \leftarrow \max_{b \in \mathcal{A}} \frac{1}{N} \sum_{i=1}^N Z_\theta(s', b, m_\phi(\tau_i))$$
            \State Compute target samples: $$z_i \leftarrow r + \gamma \cdot \mathrm{stop\_grad}(Z_\theta(s', a', m_\phi(\tau_i)))$$
        \EndFor
        \State Compute the expectile loss: $$\mathcal{L}_E\leftarrow \frac{1}{N^2}\sum_{i=1}^N \sum_{j=1}^N \left(\tau_i \mathds{1}_{z_j > e_i} + (1 - \tau_i) \mathds{1}_{z_j \leq e_i} \right) \left( z_j - e_i \right)^2 $$
        \State Compute the quantile loss: $$\mathcal{L}_{Q} \leftarrow \frac{1}{N^2}\sum_{i=1}^N \sum_{j=1}^N \left(\tau_i \mathds{1}_{z_j > q_i} + (1 - \tau_i) \mathds{1}_{z_j \leq q_i} \right) \left| z_j - q_i \right| $$
        \State Update expectile function parameters: $\theta \leftarrow \theta - \lambda \nabla_\theta \mathcal{L}_E$
        \State Update mapper parameters: $\phi \leftarrow \phi - \lambda \nabla_\phi \mathcal{L}_Q$
    \end{algorithmic}
\end{algorithm*}

The state-action embeddings of the mapper are copied form those of the $Z$-function. This way, the parameters of the $Z$-function (in our experiments below this includes the large image embedding networks and the overall scale of the rewards) are learned using expectile regression, while only the residual shape difference between the quantile and expectile function is learned by the mapper, using quantile regression. 

The update step described above can be formalized as a distributional operator, which we define in Section~\ref{convergence}. We prove that our proposed update operator converges to the distributional Bellman operator in the limit of infinite estimated quantile/expectile fractions. Then, in Section~\ref{ieqn}, we detail a practical implementation of dual expectile-quantile RL based on implicit quantile networks that we name IEQN.

\subsection{Convergence of the dual expectile-quantile operator}
\label{convergence}

In this section, we prove that our proposed update operator converges to the
distributional dynamic programming operator from Eq.~\eqref{eq:ddp} as the number of quantiles and expectiles kept in memory grows infinitely large, i.e., that the error incurred by our dual expectile-quantile operator vanishes in the limit of an infinite number of statistics to be evaluated. This result relies on several properties of the expectile function, including its absolute continuity that we establish in the following lemma:

\begin{restatable}{lemma}{absolutecontinuity}
\label{absolutecontinuity}
    Let $Z$ be a random variable taking values in $[a,b]$ with finite second moment and whose CDF admits finitely many discontinuities. Then, the expectile function $E_Z : \tau \mapsto \arg\min_{e} \mathbb{E}_{z \sim Z} [(\tau \mathds{1}_{z > e} + (1 - \tau) \mathds{1}_{z \leq e} ) ( z - e )^2 ]$ is absolutely continuous on $[0,1]$.
\end{restatable}

\noindent%
The proofs for this lemma and all results below are included in the appendix. We are now able to prove our main result, Theorem~\ref{lemma}, i.e., that our dual regression projection operator approximates the target distribution well in the limit of an infinite number of quantile/expectile fractions:

\begin{restatable}{theorem}{wassersteinbound}
\label{lemma}
Let $\tau_k = \frac{2k -1}{2K}$, for $k = 1, \dots, K$, and let $\Pi_{\mathcal{M}}^K : \mathscr{P}(\mathbb{R}) \rightarrow \mathscr{P}(\mathbb{R})$ be the dual regression projection operator defined as:

 $\forall \eta \in \mathscr{P}(\mathbb{R}),$
\begin{align}
 \Pi_{\mathcal{M}}^K(\eta) &= \frac{1}{K} \sum_{k=1}^{K} \delta_{E_\eta \left(\mathrm{floor}^K \left( E^{-1}_\eta ( F^{-1}_\eta(\tau_k) \right) \right)}  \\
 &= \frac{1}{K} \sum_{k=1}^{K} \delta_{E_\eta \left(\frac{2\left \lfloor K \mathcal{M}(\tau_k)  + 1 / 2 \right \rfloor - 1}{2K} \right)}, 
\end{align}

\noindent%
where $E_\eta : [0,1] \rightarrow \mathbb{R}$ is the expectile function of $\eta$, $F^{-1}_\eta : [0,1] \rightarrow \mathbb{R}$ is the inverse CDF -- i.e., the quantile function -- of $\eta$, and $\mathrm{floor}^K(x) = \tau_{\left \lfloor Kx  + \frac{1}{2} \right \rfloor}$. Let $\eta \in \mathscr{P}(\mathbb{R})$ be a bounded-support probability distribution with finite second moment and whose CDF admits finitely many discontinuities, and let $W_1$ be the $1$-Wasserstein distance. Then:
$$\lim_{K \to \infty} W_1(\Pi_{\mathcal{M}}^K\eta, \eta) = 0 \; .$$
\end{restatable}

\noindent%
Reusing the notation from the theorem, we can formally define our dual expectile-quantile operator. Let $\pi \in \mathscr{P}(\mathcal{A})^\mathcal{S}$ be a policy, we have:
\begin{equation}
    \mathcal{T}_{\mathcal{M}^K}^\pi = \Pi_\mathcal{M}^K \mathcal{T}^\pi \;,
\end{equation}
where $\mathcal{T}^\pi : Z(s_t,a_t) = R(s_t,a_t) + \gamma \mathbb{E}_\pi\left[Z(s_{t+1}^\pi,a_{t+1}^\pi)\right]$ is the distributional Bellman operator (see Section~\ref{sec:drl}).
We can now derive a key corollary in the context of distributional RL training:

\begin{restatable}{corollary}{convergence}
\label{corollary}
    On Markov decision processes with bounded rewards and $\gamma < 1$, the dual expectile-quantile operator converges pointwise to the distributional Bellman operator:
    $$ \lim_{K \to \infty} \mathcal{T}_{\mathcal{M}^K}^\pi = \mathcal{T}^\pi \; \mathrm{pointwise}.$$
\end{restatable}

\noindent%
This result comes in contrast to the failure of the naive expectile operator~\citep{er-dqn} to match the distributional Bellman operator. We now present a practical implementation of an agent using our dual approach.

\begin{figure*}[!ht]
\tabskip=0pt
\halign{#\cr
  \hbox{%
    \begin{subfigure}[b]{\textwidth}
    \centering
    \includegraphics[height=6.5cm, width=\textwidth]{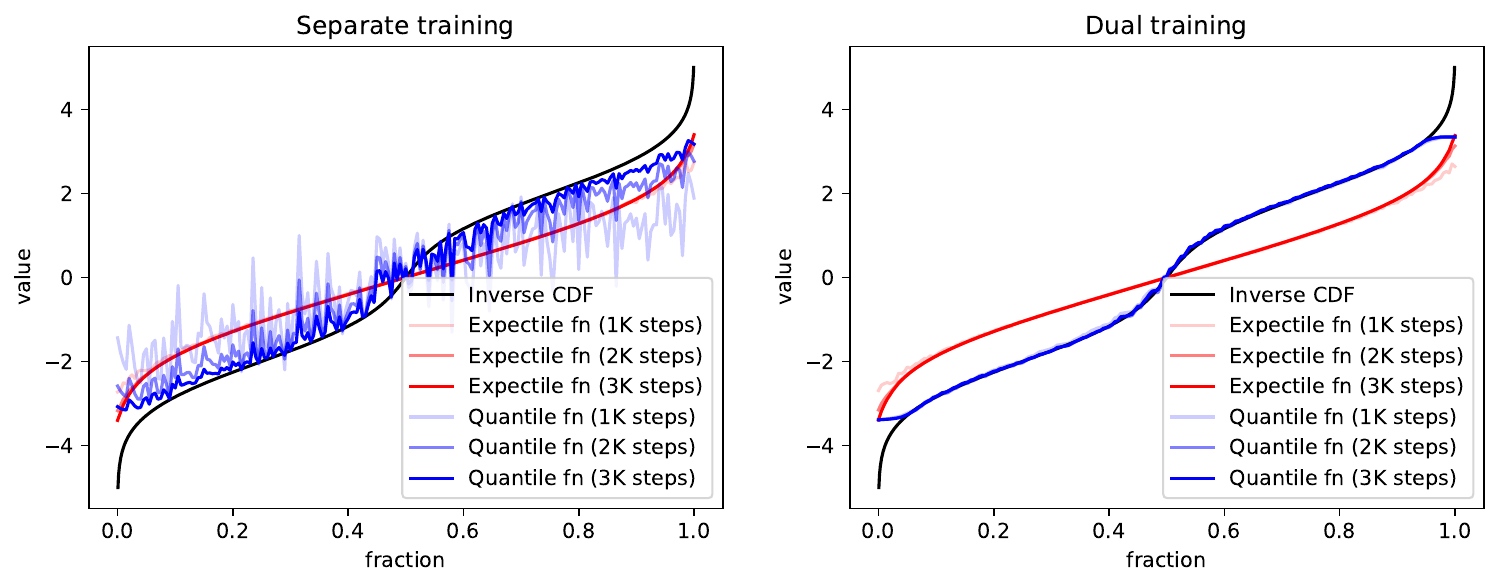}
    \caption{Approximating a distribution with separate and dual training.}
    \label{fig:separate_vs_dual_regression}
    \end{subfigure}%
  }\cr
  \hbox{%
    \begin{subfigure}{\textwidth}
    \centering
    \includegraphics[height=6.5cm, width=\textwidth]{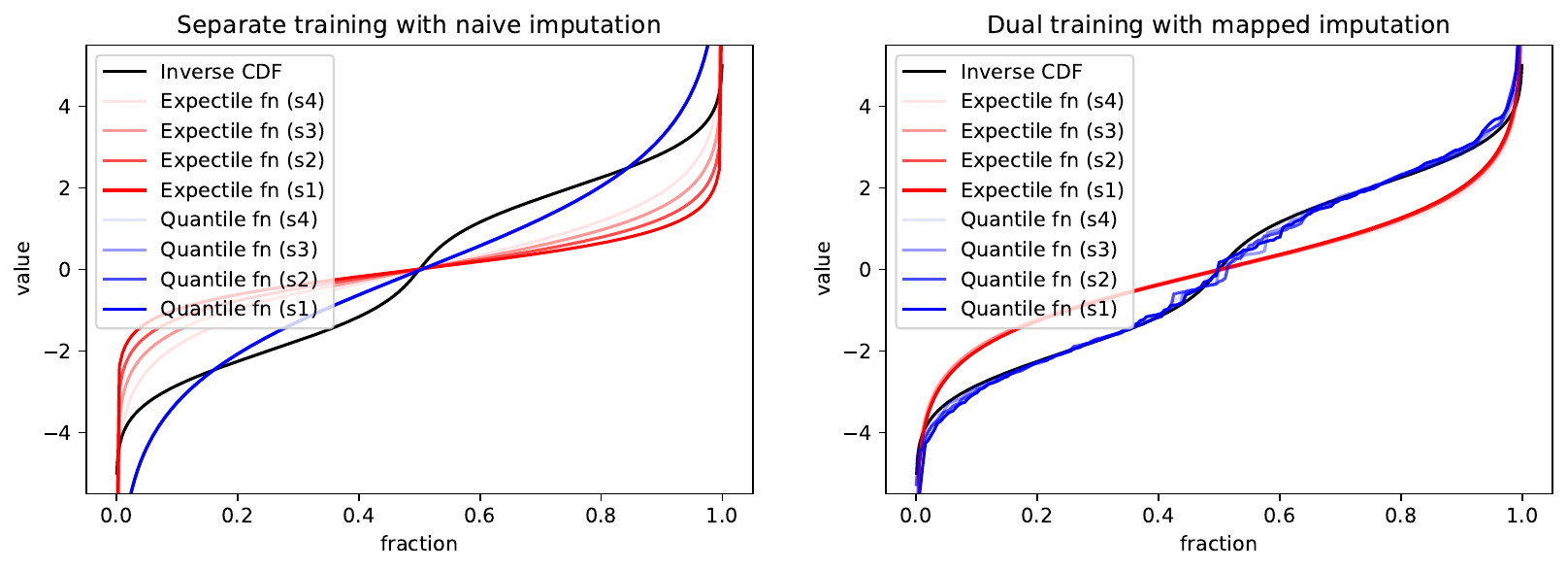}
    \caption{Tabular distributional RL with separate  and dual training.}
    \label{fig:separate_vs_dual_bellman}
    \end{subfigure}%
  }\cr
}
\caption{\textbf{(a)} Approximating a bimodal distribution with quantile and expectile regression. Quantile regression approximates the inverse CDF, albeit with high variance, especially on extreme values (left, blue curves). Expectiles converge very quickly to the expectile function (left, red curves). When training a mapper to generate quantiles from expectiles, quantile estimation becomes much more efficient (right). \textbf{(b)} Distributional RL with function approximation in a chain MDP with 4 states, and a bimodal reward distribution at the last state. The expectile function collapses as the temporal difference error propagates to previous states (left, red curves) while the quantile function is a poor approximation of the inverse CDF (left, blue curves). Our dual method solves both problems (right).}
\label{fig:separate_vs_dual}
\end{figure*}

\subsection{A practical implementation: IEQN}
\label{ieqn}
We use the principle described in Section~\ref{sec:dual-training} to implement IEQN (Algorithm~\ref{alg-ieqn}), a new distributional RL agent based on implicit quantile networks (IQN)~\citep{iqn}. The $Z$-function is modeled as a neural network inputting a state and a fraction $\tau \sim \mathcal{U}(0,1)$, and outputting $\tau$-expectile values for all actions. Its parameters are learned via an asymmetric $L_2$ loss, i.e., expectile regression. We also use a neural network to implement the mapper between quantile fractions and expectile fractions, and learn its parameters via an asymmetric $L_1$ loss, i.e., quantile regression.

\section{Experiments}
\label{sec:experiments}

We first demonstrate on a toy MDP the benefits of learning quantiles and expectiles together. We then describe our experimental setup and results on the Atari Arcade Learning Environment (ALE).

\subsection{Chain MDP: A toy example}
\label{sec:chain-MDP}

We start by observing the effect of our proposed operator in a toy environment. The MDP comprises $4$ states, each pointing to the next through a unique action and without accumulating any reward, until the last state $s_4$, where the episode terminates and the agent obtains a reward sampled from a bimodal distribution $r \sim ( \frac{1}{2}\mathcal{N}(-2, 1) + \frac{1}{2}\mathcal{N}(+2, 1) )$ (see the Appendix for a visual description).

Figure~\ref{fig:separate_vs_dual_regression} highlights the advantageous properties of expectile regression that were introduced in prior work~\citep{expectile-blue, expbible, exp-quant-david-goliath}. When trying to approximate the distribution of terminal rewards directly from samples (left), we can see that expectile regression yields more accurate estimates than quantile regression in the low-data regime (recall that the quantile function is the inverse CDF while the expectile function is in general not). Interestingly, coupling expectile regression with our mapper (right) allows us to recover the quantile function much more efficiently than quantile regression itself. We can therefore confirm the findings from prior work~\citep{expectile-blue, exp-quant-david-goliath} and conclude that our learning-based procedure for mapping estimated expectiles to their corresponding quantile fraction is effective.

In Figure~\ref{fig:separate_vs_dual_bellman}, we instantiate the problem in a typical dynamic programming setting, to illustrate the deficiencies of regular quantile and expectile dynamic programming. We can observe (left) that quantile function learning is sample-inefficient and fails to approximate the distribution within the given evaluation budget.\footnote{In this figure, we learn quantile and expectile functions parameterized by neural networks, as opposed to Figure~\ref{fig:separate_vs_dual_regression} where each statistic is learned independently from others. This explains why the quantile function's appearance is smoother in this figure.} However, the distribution information is propagated correctly through temporal difference updates, since the quantile functions estimated at each state coincide. In contrast, the expectile function collapses to the mean as the error propagates from $s_4$ to $s_1$. This is due to the fact that expectile values at the next state-action pair cannot be used as pseudo-samples of the return distribution $Z(s_{t+1})$~\citep{er-dqn}. Finally, Figure \ref{fig:separate_vs_dual_bellman}  (right) shows that our dual training method, where the pseudo-samples of $Z(s',a')$ are the estimated quantiles $Z_\theta(s_{t+1}, m_\phi(\tau))$, solves both issues: the expectile function does not collapse anymore and the quantile function approximation is an accurate estimation of the inverse CDF.

\subsection{Experiments on the Atari Arcade Learning Environment}

\subsubsection{Baselines}

We experimented with the following baselines to evaluate our approach:
\begin{description}
    \item [IQN-0, IQN-1] We approximate quantiles using the general approach described in IQN~\citep{iqn}, respectively without and with a Huber loss. 
    \item[IEN-Naive] We use a similar approach as for IQN, but trained with an expectile loss and a naive imputation step as described in~\citep{er-dqn}, i.e., expectile values are used as target for the temporal difference loss. The solver-based implementation described by the authors was too slow on our setup, as it was approximately 25 times slower than the other baselines.
\end{description}

\subsubsection{Environments}

We opted to conduct our experiments with the Atari Learning Environment (ALE)~\citep{bellemare13arcade}, following the setup of~\citet{machado18arcade}, notably including a $25\%$ chance to perform a sticky action at each step, i.e., repeating the latest action instead of using the action predicted by the agent. This creates stochasticity in the environment, which should be captured by distributional RL agents. In order to accommodate for limited computing resources, we constrained ourselves to the Atari-5 subbenchmark~\citep{atari5}, yet using 5 seeds to reduce the uncertainty in our results. We perform 25 validation episodes every $1$M steps to generate our performance curves. 
As is common with ALE, we report human-normalized scores, rather than raw game scores, and we aggregate them using the interquartile mean (IQM), as it \textit{is a better indicator of overall performance} (compared to sample median)~\citep{agarwal2021deep}, due to its robustness to scale across tasks and to outliers. It is especially needed, as the presence of sticky actions increases the number of outlier~seeds.

\subsubsection{Implementation details}

We base all baselines and our method on the same underlying neural network, implemented in JAX~\citep{jax2018github}. Its architecture follows the structure detailed by~\citet{iqn}. We used the training loop composition of CleanRL~\citep{cleanrl}. Hyperparameters can be found in the appendix. 
We implemented the $Z$-function for all agents as a feed-forward neural network with layer normalization. We did not use the fraction proposal network introduced with FQF~\citep{fqf}, as our method can be seen as complementary to it, and we focus on the effect of the choice of statistics. Finally, we found that using layer normalization increased performance for both our method and baselines. 

As described in Algorithm~\ref{alg-ieqn}, we only use the expectile loss to update the $Z$-function for our agent, while we use the quantile loss to update our mapper.
The mapper is implemented as a two layer, residual fully-connected neural network with ReLU and Tanh activations.
Since it is queried to obtain both the candidate and target values, we use a mapper-specific target network updated less frequently than the live network, using Polyak averaging~\citep{polyak}
with a weight of $0.5$. We share the parameters across all states, to simplify its architecture. We detail the implications of this choice in the appendix.

\subsubsection{Results}
\label{sec:atari}

In this section, we verify that our dual approach also provides benefits at scale, on a classic benchmark. 

\begin{figure}[t]
    \centering
    \includegraphics[clip, trim=5mm 0mm 15mm 0mm, width=\linewidth]{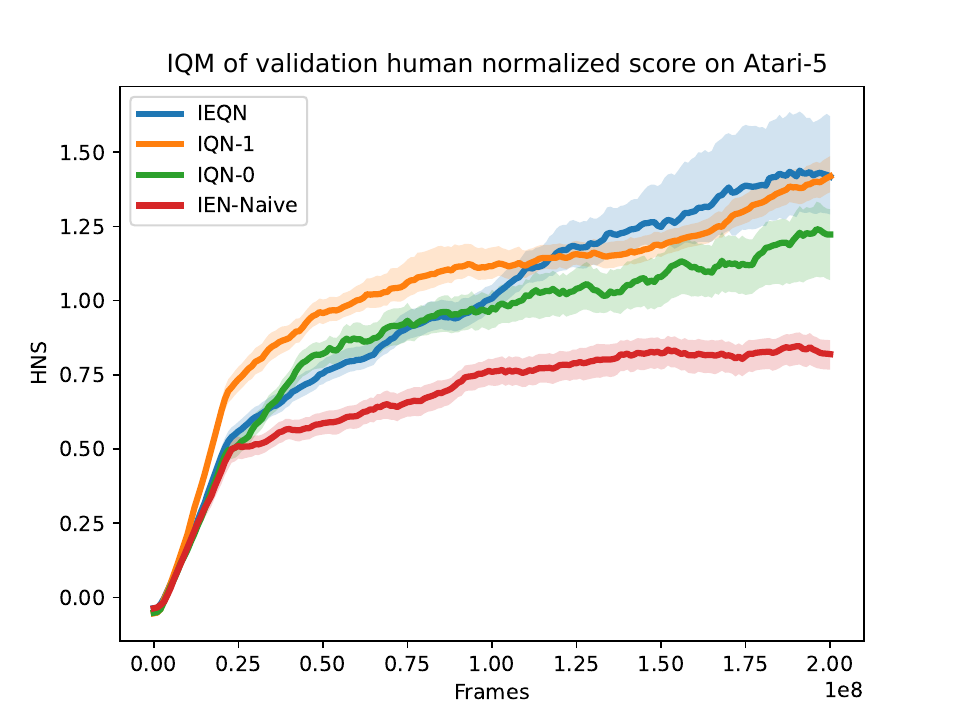}
    \caption{Interquartile mean of the human normalized score of distributional RL agents on the Atari-5 benchmark with 5 random seeds per environment. Shaded areas correspond to the 25-th and 75-th percentiles of a bootstrap distribution. A rolling average with window size of $20M$ frames is performed to enhance readability.}
    \label{fig:iqm-atari5}
\end{figure}

We first present, in Figure~\ref{fig:iqm-atari5}, the aggregated results over 5 seeds on the Atari-5 benchmark. 
We can see that despite a slower start, IEQN ends up matching the performance of IQN-1. To get statistically stronger results, we also performed a bootstrap hypothesis test on the difference of IQMs at the end of training (we average scores from the last 5 validation epochs to be robust to instabilities). We found that our method surpasses the performance of both the quantile approach (achieved significance level $0.0117$), and naive expectile approach (achieved significance level $0$), thereby demonstrating the benefits of dual regression over single regression of either quantiles or expectiles on the final performance.

Furthermore, we verify in Table~\ref{tab:distribution-spread} that IEQN avoids distributional collapse in practice. In fact, while IQN-1's estimated distribution is much narrower than IQN-0's -- a confirmation that the Huber loss causes distributional collapse, despite its better efficiency -- IEQN's quantile spread is much larger than IQN-1's. Moreover, the expectile spread of IEQN is much larger and more stable than that of IEN-Naive, suggesting that expectile distributional RL yields degenerate distributions, as noted by~\citet{er-dqn}, but that dual expectile-quantile distributional RL avoids this collapse. 

\begin{table}[t]
    \caption{Average and standard deviation of the distance between quantile (respectively expectile) $0.1$ and $0.9$, relatively to the scale of the Q-function, at the end of training.}
    \label{tab:distribution-spread}
    \centering
    \begin{tabular}{l c c}
    \toprule
       & Quantiles spread  & Expectiles spread \\
       \midrule
       IQN-0 & 1.25 $\pm$ 0.198 & -\\
       IQN-1 & 0.144 $\pm$ 0.072 & -\\
       IEN-Naive & - & 0.174 $\pm$ 0.195\\
       IEQN & 0.721 $\pm$ 0.142 & 0.465 $\pm$ 0.086\\
    \bottomrule
    \end{tabular}
\end{table} 

\section{Conclusion}
\label{sec:conclusion}

We have proposed a statistics-based approach to distributional reinforcement learning that uses the simultaneous estimation of quantiles and expectiles of the action-value distribution. Previous work only estimated quantiles or expectiles separately. Our new approach presents the advantage of leveraging the efficiency of the expectile-based loss for both expectile and quantile estimation while solving the theoretical shortcomings of expectile-based distributional reinforcement learning, which often lead to a collapse of the expectile function in practice. 

We have shown on a toy environment how the dual optimization affects the statistics recovered in distributional RL: in short, the quantile function is estimated more accurately than with vanilla quantile regression and the expectile function remains consistent after several steps of temporal difference training. We have also benchmarked our approach at scale, on the Atari-5 benchmark. Our model, IEQN, matches the performance of the Huber-based IQN-1 and surpasses that of both expectile and quantile-based agents, demonstrating its effectiveness in practical scenarios.

We open possibilities for future research to use a distributional approach that performs well and does not collapse. For future work, we plan to investigate how the dual approach can be used in risk-aware decision-making problems, and how it performs when the goal is to optimize risk metrics such as (conditional) value-at-risk. Moreover, we plan to gather insights into what type of behavior is favored by the quantile and expectile loss, respectively.

\subsubsection*{Acknowledgments}
This research was partially supported by Ahold Delhaize, through AIRLab Amsterdam, by the Dutch Research Council (NWO), under project numbers 024.004.022, NWA.1389.20.\-183, and KICH3.LTP.20.006, and by the European Union's Horizon Europe program under grant agreement No 101070212. 
All content represents the opinion of the authors, which is not necessarily shared or endorsed by their respective employers and/or sponsors.

\bibliography{main}

\newpage
\appendix 

\clearpage

\section*{Appendix}
This appendix has the following sections:
\begin{enumerate}[leftmargin=*]
    \item[\ref{appendix:hyperparameters-etc}] Hyperparameters, code and implementation details
    \item[\ref{appendix-mapper}] Sharing the mapper's parameters
    \item[\ref{toy-MDP}] Toy Markov decision process
    \item[\ref{app:proofs}] Proof of Lemma~\ref{absolutecontinuity}
    \item[\ref{appendix:proof-of-theorem-2}] Proof of Theorem~\ref{lemma}
    \item[\ref{appendix:proof-of-corollary-3}] Proof of Corollary~\ref{corollary}
    \item[\ref{appendix:proof-of-corollary-3}] Analysis of the estimated variance
    
\end{enumerate}

\section{Hyperparameters, code and implementation details}
\label{appendix:hyperparameters-etc}
\subsection{Hyperparameters}
\label{hyperparams}

We use JAX~\citep{jax2018github} to train our models.
A full training procedure of $200$M training frames and corresponding validation epochs takes approximately $50$ hours in our setup.    

\begin{table}[!ht]
    \caption{$Z$-function hyperparameters.}
    \label{tab:shared-hyperparams}
    \centering
    \begin{tabular}{l c}
        \toprule
        Key & Value \\
        \midrule
         Discount factor & $0.99$ \\
         Batch size & $32$  \\
         Fraction distribution & $\mathcal{U}([0,1])$ \\
         Learning rate & $1\mathrm{e}^{-4}$ \\
         Random frames before training & $200000$ \\
         Size of convolutional layers & $[32,64,64]$ \\
         Size of fully-connected layer & $512$ \\
         Critic updates per sample & $2$ \\
         Buffer size & $1\mathrm{e}6$ \\
         Frames between target network updates & $35000$ \\
         Target network update rate & $1.0$ \\
         \bottomrule
    \end{tabular}
\end{table}

\begin{table}[!ht]
    \caption{Mapper hyperparameters.}
    \label{tab:mapper-hyperparams}
    \centering
    \begin{tabular}{lc}
        \toprule
        Key & Value \\
        \midrule
         Layer size & $64$ \\
         Learning rate & $7\mathrm{e}^{-5}$ \\
         Target network update rate & $0.5$  \\
         \bottomrule
    \end{tabular}
\end{table}

\subsection{Code}

Our training and evaluation loop is based on CleanRL~\citep{cleanrl}. The code base is available on \url{https://github.com/samijullien/ieqn}.

\section{Sharing the mapper's parameters}
\label{appendix-mapper}

Sharing the mapper's parameters across states and actions allows us to lighten the computational burden, which is part of the goal of this paper. We found this technique to work well in practice on the Atari-5 benchmark, although it requires additional assumptions in theory. We review
these assumptions in this section.

\citet{expectile-location-scale} show that there exists such a shared mapping between quantiles and expectiles when the regression follows a location-scale model, i.e., for random variables $X$ and $Y$: 
$$Y = \mu(X) + \sigma(X) \varepsilon,$$
where $\mu$ and $\sigma$ are continuous functions, $\varepsilon$ is centered and finite-variance, and $\varepsilon, X$ are independent. When the return distribution follows this model, $X$ being the state-action variable in this context, sharing the mapper's parameters is theoretically valid. While this may seem limiting, it does not require all state-action pairs to be allocated the same distributions, only that they share a common shape. Moreover, the location-scale family is quite broad, as it includes, e.g., Normal, Student, Cauchy, GEV distributions, and more~\citep{WeiEA2014}. 

In many distributional reinforcement learning scenarios, the assumption may be satisfied. For instance, when the environmental stochasticity emerges from small, independent perturbations, i.e., normally-distributed errors, the return distribution at every state will still be normally distributed as convolutions of Gaussian distributions are also Gaussian. On the other hand, this assumption can fail under high-frequency transition distributions, i.e., branching behaviors, where the same state-action pair can yield drastically different outcomes and the reward-next-state distribution has non-continuous support. We leave for future work the investigation of when sharing the mapper's parameters across state-action pairs fails in practice.

\section{Toy Markov decision process}
\label{toy-MDP}

\begin{figure}[!h]
    \includegraphics[width=.9\columnwidth]{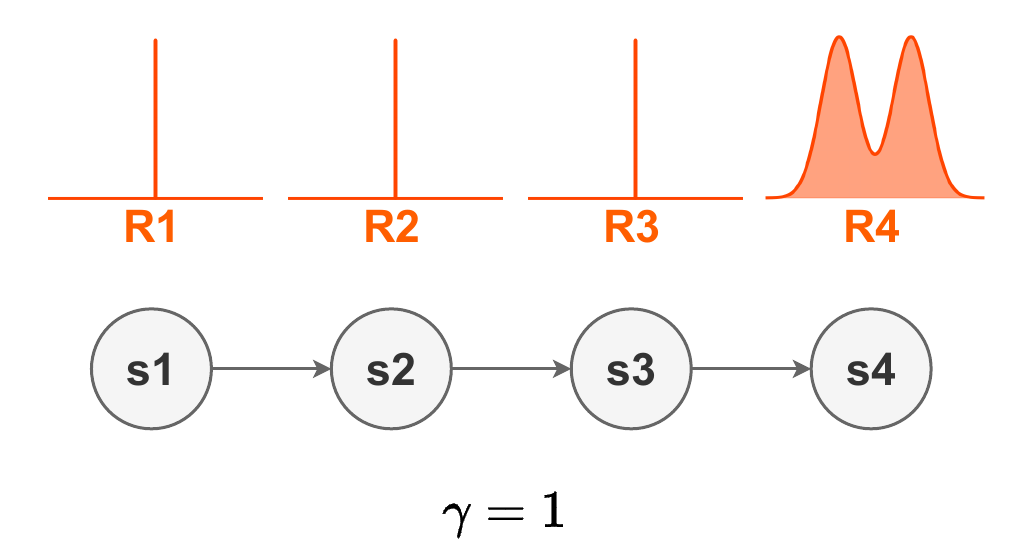}
    \caption{Toy Markov decision process.}
    \label{fig:toy-mdp}
\end{figure}

\begin{figure*}[!ht]

\section{Proof of Lemma~\ref{absolutecontinuity}}
\label{app:proofs}

    Our proof of Theorem~\ref{lemma} requires the absolute continuity of the expectile function. Therefore, we first prove the following lemma:

\absolutecontinuity*

\begin{proof}
Our proof relies on the Banach-Zarecki theorem~\citep{zaretsky}, which states that any real-valued function $f$ defined on a real bounded closed interval is absolutely continuous if and only if on this interval:
\begin{enumerate}[label=(\roman*)]
    \item $f$ is continuous;
    \item $f$ has bounded variation; and
    \item $f$ follows the Luzin N property~\citep{luzin}, i.e., the image by $f$ of a set with null Lebesgue measure also has null Lebesgue measure.
\end{enumerate}

It is well-known that the expectile function is continuous on $[0,1]$~\citep{german-paper, expbible}. Therefore, (i) is satisfied.

$E_Z$ is monotonically increasing and takes values in the finite support of $Z$. Therefore it has bounded variation and (ii) is satisfied.

In order to prove (iii), we first note that any function that is differentiable on a co-countable set has the Luzin N property~\citep{luzin}. We therefore use our assumption that $Z$ admits a finite number of discontinuities in the following.

Let $F_Z$ be the CDF of $Z$ and $D = \left\{ z \in [a,b] : \lim_{x \to z} F_Z(x) \neq F_Z(z) \right\}$ be the finite set of points at which $F_Z$ is not continuous. $D$ is a finite set within a metric space and therefore closed. As a consequence, its complement $C_{[a,b]} = [a,b] \setminus D$ is open in $[a,b]$, i.e., $\forall z \in C_{[a,b]}, \exists \varepsilon > 0$ such that $\forall x \in [a,b] d(x,z) < \varepsilon \Rightarrow x\in C_{[a,b]}$. In other words, if $F_Z$ is continuous at a point within $[a,b]$, it is also continuous in a neighborhood of that point within $[a,b]$. By assumption, the set $C^\mathcal{N}_{[a,b]} = \left\{ z \in [a,b] : \exists \varepsilon > 0, \forall x \in [a,b] , d(x,z) < \varepsilon \Rightarrow x\in C_{[a,b]}\right\}$ of points where $F_Z$ is continuous in a neighborhood of said point is therefore co-finite.

It has been shown that the expectile function $E_Z$ is continuously differentiable at any point $\tau \in [0,1]$ such that $F_Z$ is continuous in a neighborhood of $E_Z(\tau)$~\citep{german-paper, expectilesoriginal}. The expectile function is bijective~\citep{expbible} so the set of points where $E_Z$ is differentiable $\mathcal{D}_{[a,b]}^{E_Z} = E_Z^{-1}\left(C^\mathcal{N}_{[a,b]} \right)$ is also a co-finite set.

The expectile function is differentiable on a co-finite (and thus co-countable) set, i.e., it has the Luzin N property~\citep{luzin}, which yields (iii). 

We can finally apply the Banach-Zarecki theorem and conclude that the expectile function $E_Z$ is absolutely continuous on $[0,1]$.
\end{proof}
\end{figure*}

\begin{figure*}[t]
\section{Proof of Theorem~\ref{lemma}}
\label{appendix:proof-of-theorem-2}
We can now use the absolute continuity of the expectile function under our assumptions to prove the following theorem:

\wassersteinbound*

\begin{proof}
Thanks to the triangle inequality, we have :
\begin{equation} 
\begin{split}
        W_1(\Pi_{\mathcal{M}}^K\eta, \eta) \leqslant W_1(\Pi_{\mathcal{M}}^K\eta, \Pi_Q^K\eta) + W_1(\Pi_Q^K\eta, \eta) \;, 
\end{split}
\end{equation}
where $\Pi_Q^K$ is the projected quantile regression estimator defined as:
$$
\forall \eta \in \mathscr{P}(\mathbb{R}), \;\; \Pi_Q^K(\eta) = \frac{1}{K} \sum_{k=1}^{K} \delta_{F^{-1}_\eta(\tau_k)} \;.
$$
\citet[Lemma 3.2]{er-dqn} showed that $W_1(\Pi_Q^K\eta, \eta) = \mathcal{O}\left( \frac{1}{K} \right)$. We now turn to the first term:
\begin{align}  
        W_1(\Pi_{\mathcal{M}}^K\eta, \Pi_Q\eta) &= \sum_{i=0}^{K-1} \frac{1}{K} \left| E_\eta \left(\mathrm{floor}^K \left( E^{-1}_\eta \left( F^{-1}_\eta \left(\frac{2i + 1}{2K} \right) \right)  \right)\right) - F^{-1}_{\eta} \left( \frac{2i + 1}{2K}\right) \right| 
        \\
        & = \sum_{i=0}^{K-1} \frac{1}{K} \left| E_\eta \left(\mathrm{floor}^K \left( E^{-1}_\eta \left( F^{-1}_\eta \left(\frac{2i + 1}{2K} \right) \right)  \right)\right) - 
        E_\eta \left( E^{-1}_\eta \left(F^{-1}_{\eta} \left( \frac{2i + 1}{2K}\right)\right)\right) \right| 
        \\
        & \leqslant \sum_{i=0}^{K-1} \frac{1}{K} \left| E_\eta \left(\mathrm{floor}^K \left( E^{-1}_\eta \left( F^{-1}_\eta \left(\frac{2i + 1}{2K} \right) \right)  \right)\right) -
        E_\eta \left( \mathrm{floor}^K \left( E^{-1}_\eta \left( F^{-1}_\eta \left(\frac{2i + 1}{2K} \right) \right) \right) + \frac{1}{K}\right) \right|,
\end{align}
where the last inequality is obtained thanks to the monotonicity of the expectile function. By absolute continuity of the expectile function under our assumptions (proven in Lemma~\ref{absolutecontinuity}), we have:
\begin{align}  
    & \lim_{K \to \infty} \left| E_\eta \left(\mathrm{floor}^K \left( E^{-1}_\eta \left( F^{-1}_\eta \left(\frac{2i + 1}{2K} \right) \right)  \right)\right) - 
    E_\eta \left( \mathrm{floor}^K \left( E^{-1}_\eta \left( F^{-1}_\eta \left(\frac{2i + 1}{2K} \right) \right) \right) + \frac{1}{K}\right) \right| = 0,
\end{align}
from which we can deduce $\lim_{K \to \infty} W_1(\Pi_{\mathcal{M}}^K\eta, \Pi_Q\eta)= 0$ and finally $\lim_{K \to \infty} 
 W_1(\Pi_{\mathcal{M}}^K\eta, \eta) = 0$.
\end{proof}
\end{figure*}

\begin{figure*}[!ht]
\section{Proof of Corollary~\ref{corollary}}
\label{appendix:proof-of-corollary-3}

Finally, we can derive our main result for the use of distributional dynamic programming with both quantiles and expectiles:

\convergence*

\begin{proof}
    We have $\mathcal{T}_{\mathcal{M}^K}^\pi = \Pi_\mathcal{M}^K \mathcal{T}^\pi$.
    \citet{distributional-book} have shown that the set of empirical distributions $\mathcal{F}_E$ is closed under the operator $\mathcal{T}^\pi$ (Proposition~5.7). Thus, for any empirical return distribution $\eta \in \mathcal{F}_E$, $\mathcal{T}^\pi \eta$ is also empirical and its CDF admits finitely many discontinuities. Moreover, it has bounded support. Indeed, if, without loss of generality, we consider that the reward distribution take values in $[0, R_\mathrm{max}]$, we have that every possible return distribution $\eta$ takes values in $[0, \frac{R_\mathrm{max}}{1-\gamma} ]$, and therefore $\mathcal{T}^\pi \eta$ takes values in $[0, R_{max} + \gamma \frac{R_\mathrm{max}}{1-\gamma} ] = [0, \frac{R_\mathrm{max}}{1-\gamma} ]$.

    We can now apply Theorem~\ref{lemma}:
    $$ \forall \eta \in \mathcal{F}_E \;,  \lim_{K \to \infty} 
W_1(\Pi_{\mathcal{M}}^K\mathcal{T}^\pi \eta, \mathcal{T}^\pi \eta) = 0, $$ 
and the result immediately follows. 
\end{proof}
\end{figure*}

\begin{figure*}[!ht]
\section{Analysis of the estimated variance}

In this section, we perform an additional experiment to better assess the quality of the value distribution on the Atari task. The distribution learned in our method as well as all baselines estimates the optimal Z-function, i.e., the return distribution of the optimal policy, which we cannot have ground truth for on large-scale tasks. We may however assume that the greedy policy gets closer to the optimal policy towards the end of training. If we do so, then we can compare the learned Z-function with the return distribution obtained by unfolding our agent's policy. Below, we show the variance of the learned Z-function (Figure \ref{fig:pred_variance}), and the average deviation between this prediction and the observed squared differences when rolling out the policy (Figure \ref{fig:mae_variance}), throughout the first 50M steps of training on Battlezone.

\begin{subfigure}[b]{\columnwidth}
    \includegraphics[width=\columnwidth]{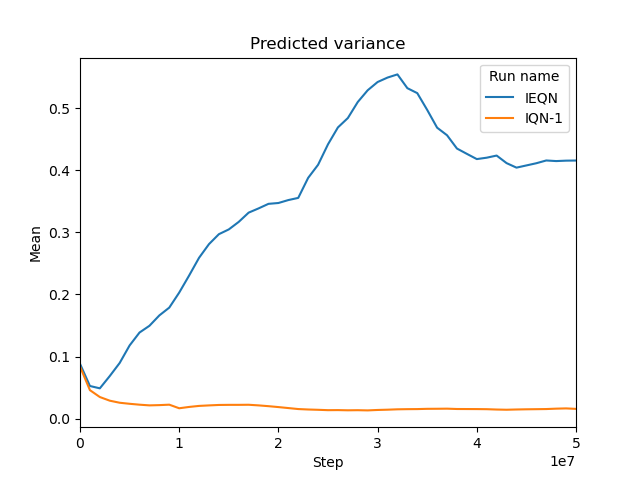}
    \caption{Predicted variance on Battlezone during training.}
    \label{fig:pred_variance}
\end{subfigure}%
\begin{subfigure}[b]{\columnwidth}
    \includegraphics[width=\columnwidth]{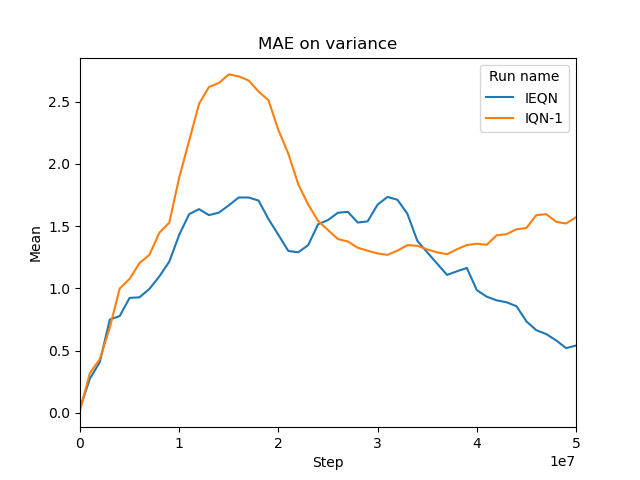}
    \caption{Error on variance on Battlezone during training.}
    \label{fig:mae_variance}
\end{subfigure}%
\label{fig:variance_analysis}
\caption{Comparison of estimated variance against observed variance of unfolding the greedy policy.}

\bigskip

We can see that (i) IQN-1 predicts a very low variance compared to IEQN, and (ii) using the approximation that the current policy is close to the optimal policy, IEQN's prediction gets closer to the observed variance than IQN-1's, as training progresses. 
\end{figure*}
\end{document}